\documentclass{article}

\usepackage[preprint,nonatbib]{neurips_2019}
\usepackage[numbers]{natbib}

\usepackage[utf8]{inputenc} 
\usepackage[T1]{fontenc}    
\usepackage{hyperref}       
\usepackage{url}            
\usepackage{booktabs}       
\usepackage{amsfonts}       
\usepackage{nicefrac}       
\usepackage{microtype}      
\usepackage{ulem}          
\usepackage{color}

\usepackage{wrapfig}

\usepackage{subcaption}
\usepackage{multirow}

\usepackage{amsmath,amsfonts,bm}

\def\eqref#1{equation~\ref{#1}}

\def\1{\bm{1}}

\DeclareMathAlphabet{\mathsfit}{\encodingdefault}{\sfdefault}{m}{sl}
\SetMathAlphabet{\mathsfit}{bold}{\encodingdefault}{\sfdefault}{bx}{n}

\usepackage{todonotes}

\title{GraphNVP: An Invertible Flow Model for Generating Molecular Graphs}

\author{%
  Kaushalya Madhawa\thanks{Work done during his stay at Preferred Networks, Inc.} \\
  Tokyo Institute of Technology\\
  Tokyo, Japan\
  \texttt{kaushalya@net.c.titech.ac.jp} \\
  \And
  Katushiko Ishiguro~~~~~Kosuke Nakago~~~~~Motoki Abe \\
  Preferred Networks, Inc.\\
  Tokyo, Japan\\
  \texttt{k.ishiguro.jp@ieee.org, \{ishiguro, nakago, motoki\}@preferrred.jp} \\
}

\begin{document}

\maketitle

\begin{abstract}
We propose GraphNVP, the first invertible, normalizing flow-based molecular graph generation model. We decompose the generation of a graph into two steps: generation of (i) an adjacency tensor and (ii) node attributes. 
This decomposition yields the exact likelihood maximization on graph-structured data, combined with two novel reversible flows. 
We empirically demonstrate that our model efficiently generates valid molecular graphs with almost no duplicated molecules. In addition, we observe that the learned latent space can be used to generate molecules with desired chemical properties.
\end{abstract}

\section{Introduction}

Generation of molecules with certain desirable properties is a crucial problem in computational drug discovery. Recently, deep learning approaches are being actively studied for generating promising candidate molecules quickly. 
Earlier models~\cite{kusner2017gvae, gomez2018automatic} depend on a string-based representation of molecules. 
However, recent models~\cite{jtvae2018,gcpn2018,de2018molgan} directly work on molecular graph representations and record impressive experimental results.  
In these studies, either variational autoencoder (VAE)~\cite{kingma2013vae} or generative adversarial network (GAN)~\cite{goodfellow2014gan, dcgan2015} are used mainly to learn mappings between the graphs and their latent vector representations. 

In this paper, we propose \textit{GraphNVP}, yet another framework for molecular graph generation based on the invertible normalizing flow~\cite{nice2015, realnvp2017, glow2018}, which was mainly adopted for image generation tasks. 
However, the sparse and discrete structure of molecular graphs is quite different from the regular grid image pixels. To capture distributions of such molecular graphs into a latent representation, we propose a novel two-step generation scheme. Specifically, GraphNVP is equipped with two latent representations for a molecular graph: first for the graph structure represented by an adjacency tensor, and second for node (atom) label assignments. During the generation process, Graph NVP first generates a graph structure. Then node attributes are generated according to the structure. This two-step generation enables us to generate valid molecular graphs efficiently.

A significant advantage of the invertible flow-based models is they perform precise likelihood maximization, unlike VAEs or GANs. We believe precise optimization is crucial in molecule generation for drugs, which are highly sensitive to a minor replacement of a single atom (node). 
For that purpose, we introduce two types of reversible flows that work for the aforementioned two latent representations. 

Thanks to the reversibility of the flow models, new graph samples can be generated by simply feeding a latent vector into the same model but in the reverse order. 
In contrast, VAE models are made of a stochastic encoder and an imperfect decoder. The decoder learns to generate a sample from a given latent vector by minimizing a reconstruction loss. For graph data, calculating the reconstruction loss often involves computationally demanding graph matching~\cite{simonovsky2018graphvae}. VAE-based molecule generation models \cite{kusner2017gvae} sidestep the stochasticity of the decoder by decoding the same latent vector multiple times and choosing the most common molecule as the output. Since flow-based models are invertible by design, perfect reconstruction is guaranteed and no time-consuming procedures are needed. Moreover, the lack of an encoder in GAN models makes it challenging to manipulate the sample generation. For example, it is not straightforward to use a GAN model to generate graph samples that are similar to a query graph (e.g., lead optimization for drug discovery), while it is easy for flow-based models.

In the experiments, we compare the proposed flow model with several existing graph generation models using two popular molecular datasets. Surprisingly, the proposed flow model generates molecular graphs with almost 100\% uniqueness ratio: namely, the results contain almost no duplicated molecular graphs. Additionally, we show that the learned latent space can be utilized to generate molecular graphs with desired chemical properties, even though we do not encode domain expert knowledge into the model. 

\section{Related Work}

\subsection{Molecular Graph Generation}
We can classify the existing molecular graph generation models based on how the data distribution is learned. Most current models belong to two categories. First, VAE-based models assume a simple variational distribution for latent representation vectors~\cite{jtvae2018,cgvae2018,ma2018constrained}. Second, some models implicitly learn the empirical distribution, especially based on the GAN architecture (e.g.,~\cite{de2018molgan,gcpn2018,Guimaraes17ORGAN}).
Some may resort to reinforcement learning~\cite{gcpn2018} to alleviate the difficulty of direct optimization of the objective function. We also observe an application of autoregressive recurrent neural networks (RNN) for graphs~\cite{You18graphrnn}. 
In this paper, we will add a new category to this list: namely, the invertible flow.

Additionally, we can classify the existing models based on the process they use for generating a graph. There are mainly two choices in the generation process. 
One is a sequential \textit{iterative} process, which generates a molecule in a step-by-step fashion by adding nodes and edges one by one~\cite{jtvae2018, gcpn2018}. 
The alternative is \textit{one-shot} generation of molecular graphs, when the graph is generated in a single step. This process resembles commonly used image generation models (e.g.,~\cite{glow2018}).
The former process is advantageous in (i) dealing with large molecules and (ii) forcing validity constraints on the graph (e.g., a valency condition of molecule atoms). 
The latter approach has a major advantage: the model is simple to formulate and implement. This is because the one-shot approach does not have to consider arbitrary permutations of the sequential steps, which can grow exponentially with the number of nodes in the graph. 

Combining these two types of classification, we summarize the current status of molecular graph generation in Table~\ref{tab:model_summary}. 
In this paper, we propose the first graph generation model based on the invertible flow, with one-shot generation strategy. 

\begin{table}[t]
    \centering
    \begin{tabular}{c||c|c|c|c|c||c|c}
        \multirow{2}{*}{Name} & \multicolumn{5}{c||}{Distribution Model} & \multicolumn{2}{c}{Generation Process} \\
         & VAE & Adversarial & RL & RNN & InvertibleFlow & Iterative & One-shot\\ \hline
        RVAE~\cite{ma2018constrained} & \checkmark & - & - & - & - & -  & \checkmark \\
        CGVAE~\cite{cgvae2018} & \checkmark & - & - & - & - & \checkmark & - \\
        JT-VAE~\cite{jtvae2018} & \checkmark & - & - & - & - & \checkmark & - \\
        MolGAN~\cite{de2018molgan} & - & \checkmark & - & - & - & - & \checkmark \\
        GCPN~\cite{gcpn2018} & - & \checkmark & \checkmark & - & - & \checkmark & - \\ 
        GraphRNN~\cite{You18graphrnn} & - & - & - & \checkmark & - & \checkmark & - \\ \hline
        \textbf{GraphNVP} & - & - & - & - & \checkmark & - & \checkmark
    \end{tabular}
    \vspace*{2mm}
    \caption{Existing models of molecular graph generation. We propose the first invertible flow-based graph generation model in the literature..}
    \label{tab:model_summary}
\end{table}

\subsection{Invertible Flow Models}

To the best of our knowledge, the normalizing flow (which is a typical invertible flow) was first introduced to the machine learning community by \cite{Tabak_VandenEijnden10,Tabak_Turner13}. Later, Rezende et al.~\cite{Rezende_Mohamed15ICML} and Dinh et al.~\cite{nice2015} leveraged deep neural networks in defining tractable invertible flows. Dinh et al. ~\cite{nice2015} introduced reversible transformations for which the log-determinant calculation is tractable. These transformations, known as \textit{coupling layers}, serve as the basis of recent flow-based image generation models ~\cite{realnvp2017,glow2018,ffjord}.
Recently, Behrman et al.~\cite{iresnet2019} introduced an invertible ResNet architecture based on numerical inversion computations. 

So far, the application of flow-based models is mostly limited to the image domain. As an exception, Kumar et al.~\cite{Kumar18grevnet} proposed flow-based invertible transformations on graphs. However, their model is only capable of modeling the node assignments and it is used for improving the performance of node and graph classification. Their model cannot learn a latent representation of the adjacency tensor; therefore, it cannot generate graph structures. We overcome this issue by introducing two latent representations, one for node assignments and another for the adjacency tensor, to capture the unknown distributions of the graph structure and its node assignments. 
Thus, we consider our proposed model to be the first invertible flow model that can generate attributed graphs.

\section{GraphNVP: Flow-based graph generation model}

\subsection{Formulation}
We use the notation $G = (A, X)$ to represent a graph $G$ consisting of an adjacency tensor $A$ and a feature matrix $X$.
Let there be $N$ nodes in the graph. Let $M$ be the number of types of nodes and $R$ be the number of types of edges. Then $A \in \{0, 1\}^{N\!\times\!N\!\times\!R}$ and $X \in \{0, 1\}^{N\!\times\!M}$. In the case of molecular graphs, $G = (A, X)$ represents a molecule with $R$ types of bonds (single, double, etc.) and $M$ types of atoms (e.g., oxygen, carbon, etc.). Our objective is to learn an invertible model $f_{\theta}$ with parameters $\theta$ that maps $G$ into a latent point $z = f_{\theta}(G) \in \mathbb{R}^{D=(N\!\times\!N\!\times\!R)+(N\!\times\!M)}$. We describe $f_{\theta}$ as a normalizing flow composed of multiple invertible functions. 

Let $z$ be a latent vector drawn from a known prior distribution $p_z(z)$ (e.g., Gaussian): $z \sim p_z(z)$. With the change of variable formula, the log probability of a given graph $G$ can be calculated as: 
\begin{equation}
\log \left(p_{G}(G)\right)=\log \left(p_{z}(z)\right)+\log \left(\left|\operatorname{det}\left(\frac{\partial z}{\partial G} \right)\right|\right),
\label{eqn:log_likelihood}
\end{equation}
where $\frac{\partial z}{\partial G}$ is the Jacobian of $f_{\theta}$ at $G$.

\subsection{Graph Representation}

Directly applying a continuous density model on discrete components may result in degenerate probability distributions. Therefore, we cannot directly employ the change of variable formula (Eq.~\ref{eqn:log_likelihood}) for these components. It is a standard practice to convert the discrete data distribution into a continuous distribution by adding real-valued noise \cite{theis2016note}. This process is known as \textit{dequantization}. Usually, dequantization is performed by adding uniform noise to the discrete data \cite{realnvp2017,glow2018}. We follow this process by adding uniform noise to components of $G$; $A$, and $X$, as shown in Eq. \ref{eqn:A_deq} and Eq. \ref{eqn:X_deq}. The dequantized graph denoted as $G' = (A', X')$ is used as the input in Eq.~\ref{eqn:log_likelihood}: 

\begin{equation}
    A' = A + c u;~ u \sim U[0, 1)^{N \times N \times R} \, ,
    \label{eqn:A_deq}
\end{equation}
\begin{equation}
    X' = X + c u;~ u \sim U[0, 1)^{N \times M} \, ,
    \label{eqn:X_deq}
\end{equation}
where $0 < c < 1$ is a scaling hyperparameter. We adopted $c = 0.9$ for our experiment.

Note that the original discrete inputs $A$ and $X$ can be recovered by simply applying floor operation on each continuous value in $A'$ and $X'$. Hereafter, all the transformations are performed on dequantized inputs $A'$ and $X'$.

\begin{figure}[t]
    \centering
    \includegraphics[width=0.9\linewidth]{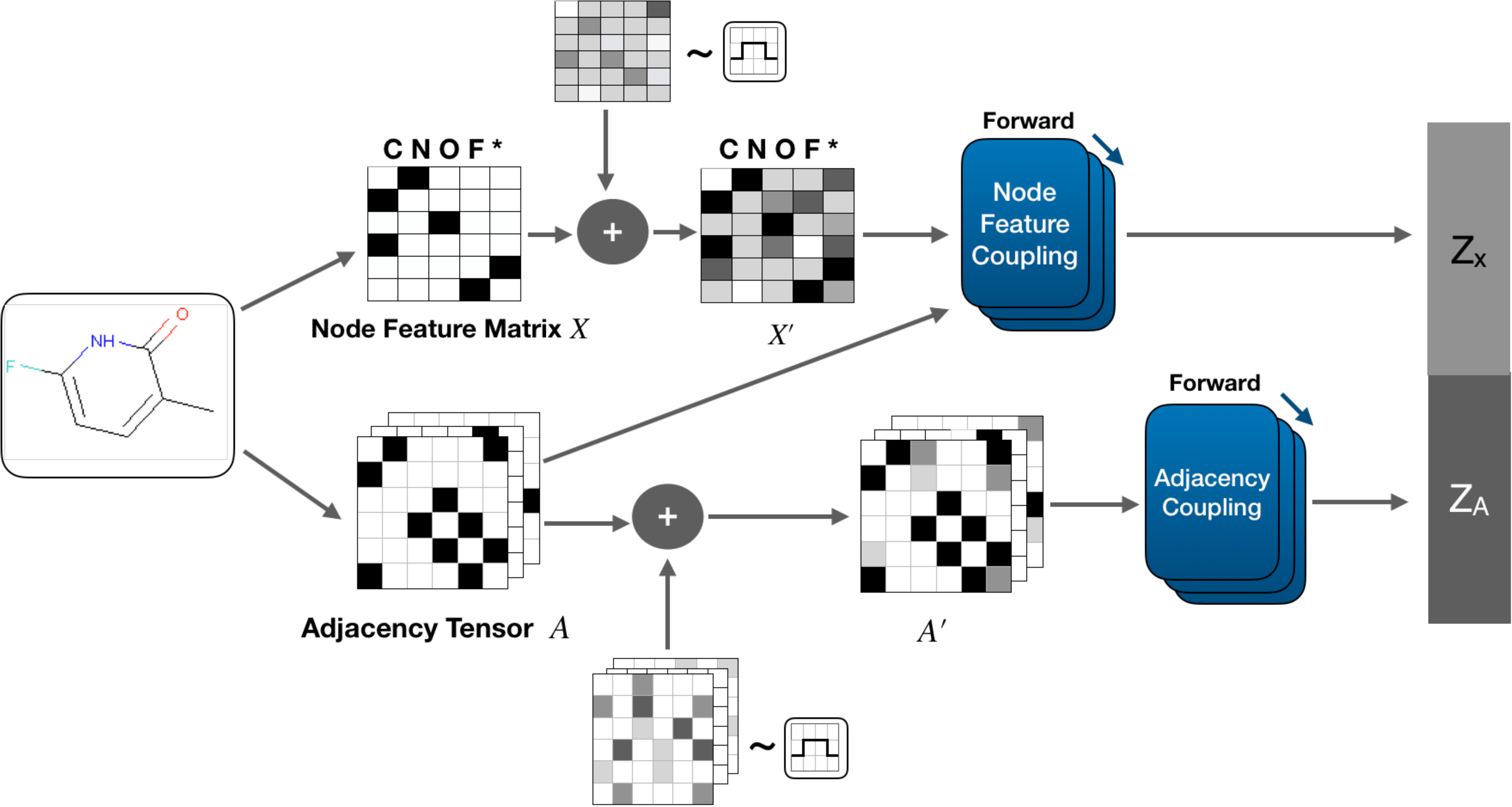}
    \caption{Forward transformation of the proposed GraphNVP.}
    \label{fig:model_forward}
\end{figure}

\subsection{Coupling layers}

Based on real-valued non-volume preserving (real NVP) transformations introduced in ~\cite{realnvp2017}, we propose two types of reversible affine coupling layers; \textit{adjacency coupling} layers and \textit{node feature coupling} layers that transform the adjacency tensor $A'$ and the feature matrix $X'$ into latent representations, $z_A \in \mathbb{R}^{N\!\times\!N\!\times\!R}$ and $z_X \in \mathbb{R}^{N\!\times\!M}$, 
respectively. 


We apply $L_X$ layers of node feature coupling layers to a feature matrix $X'$ to obtain $z_X$. We denote an intermediate representation of the feature matrix after applying the $\ell$\textsuperscript{th} node feature coupling  layer as $z_{X}^{(\ell)}$. Starting from $z_X^{(0)} = X'$, we repeat updating rows of $z_X$ over $L_X$ layers. 
Each row of $z_X^{(\ell)}$ corresponds to a feature vector of a node in the graph. Finally, we obtain $z_X = z_X^{(L_X)}$ as the final latent representation of the feature matrix. The $\ell$\textsuperscript{th} node feature coupling layer updates a single row $\ell$ of the feature matrix while keeping the rest of the input intact: 
\begin{equation}
	z_X^{(\ell)} [\ell,:] \leftarrow z_X^{(\ell-1)} [\ell, :] \odot \exp \left( s(z_X^{(\ell-1)} [\ell^{-}, :], A) \right)+ t(z_X^{(\ell-1)} [\ell^{-}, :], A),
	\label{eqn:node_feature_coupling}
\end{equation}
where functions $s$ and $t$ stand for scale and translation operations, and $\odot$ denotes element-wise multiplication. We use $z_X[\ell^{-}, :]$ to denote a latent representation matrix of $X'$ excluding the $\ell$\textsuperscript{th} row (node). 
Rest of the rows of the feature matrix will stay the same as
\begin{equation}
    z_X^{(\ell)} [\ell^{-},:] \leftarrow z_X^{(\ell-1)} [\ell^{-},:].
\end{equation}{}
Both $s$ and $t$ can be formulated with arbitrary nonlinear functions, as the reverse step of the model does not require inverting these functions. Therefore, we use the graph adjacency tensor $A$ when computing invertible transformations of the node feature matrix $X'$. So as functions $s$ and $t$ in a node feature coupling layer, we use a sequence of generic graph neural networks. It should be noted that we use the discrete adjacency tensor $A$, as only the node feature matrix is updated in this step. In this paper, we use a variant of Relational GCN~\cite{relgcn2018} architecture.  

Likewise, we apply $L_A$ layers of transformations for the adjacency tensor $A'$ to obtain the latent representation $z_A$. We denote an intermediate representation of the adjacency tensor after applying the $\ell$\textsuperscript{th} adjacency coupling as $z_{A}^{(\ell)}$. The $\ell$\textsuperscript{th} adjacency coupling layer updates only a single slice of $z_{A}^{\ell}$ with dimensions $N\!\times\!R$ as: 
\begin{equation}
z_A^{(\ell)} [\ell,:, :] \leftarrow z_A^{(\ell-1)} [\ell, :, :] \odot \exp \left( s(z_A^{(\ell-1)} [\ell^{-}, :, :]) \right) + t(z_A^{(\ell-1)} [\ell^{-}, :, :]).
\label{eqn:adjacency_coupling}
\end{equation}
The rest of the rows will stay as it is:
\begin{equation}
    z_A^{(\ell)} [\ell^{-},:, :] \leftarrow z_A^{(\ell-1)} [\ell^{-}, :, :].
\end{equation}
For the adjacency coupling layer, we adopt multi-layer perceptrons (MLPs) for $s$ and $t$ functions. Starting from $z_A^{(0)} = A'$, we repeat updating the first axis slices of $z_A$ over $L_A$ layers. Finally, we obtain $z_A = z_A^{(L_A)}$ as the final latent representation of the adjacency tensor. 

\subsubsection{Masking Patterns and Permutation over Nodes}

\begin{figure}[t]
	\centering
	\vspace{-0cm}
	\begin{subfigure}[b]{0.45\textwidth}
		\begin{center}
			\centering
			\includegraphics[width=.80\textwidth]{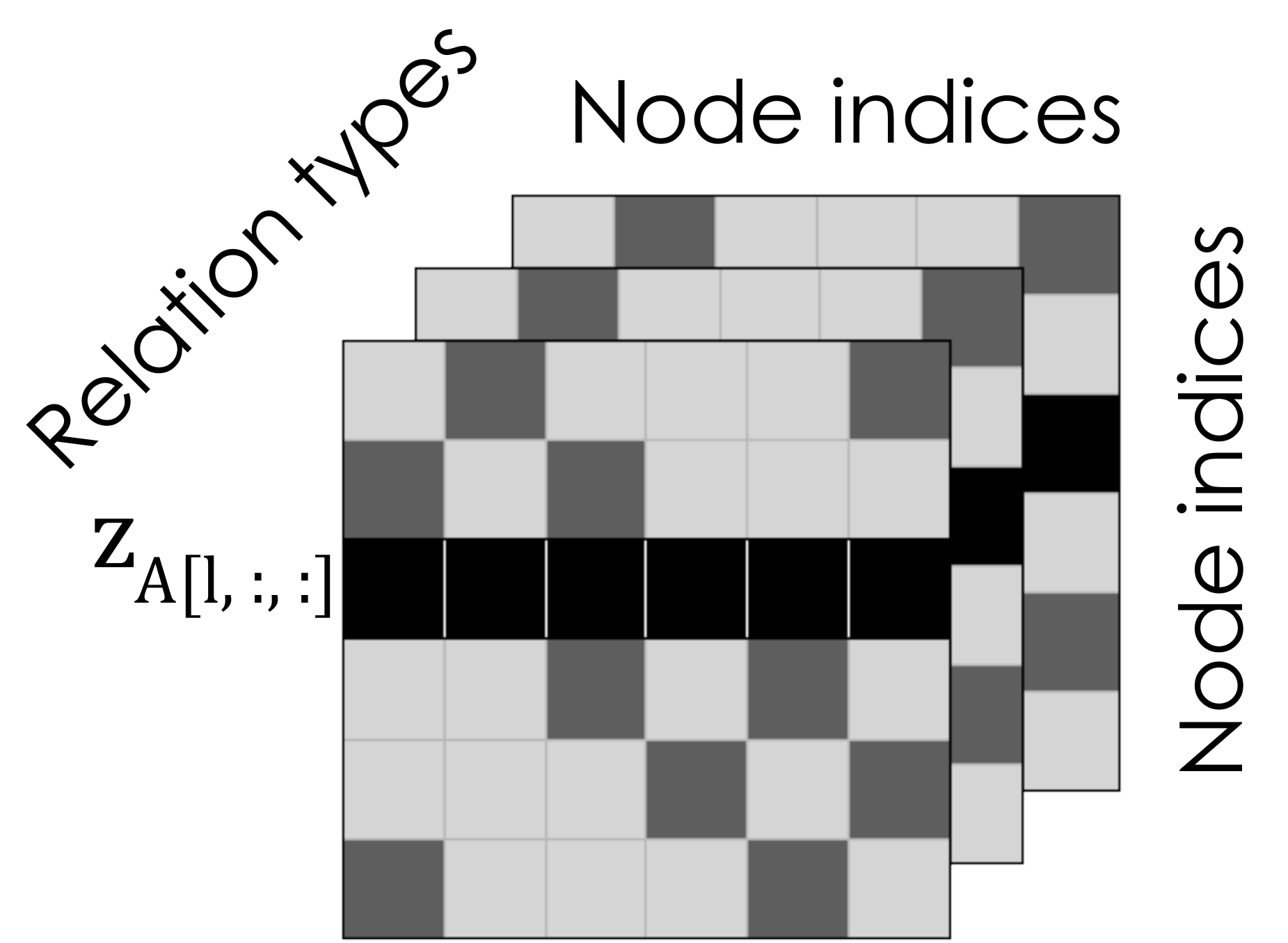}
			\caption{Adjacency coupling layer: A single row of adjacency tensor is masked.}
		\end{center}
	\end{subfigure}%
	\hspace{3mm}
	\begin{subfigure}[b]{0.45\textwidth}
		\centering
		\includegraphics[width=.65\textwidth]{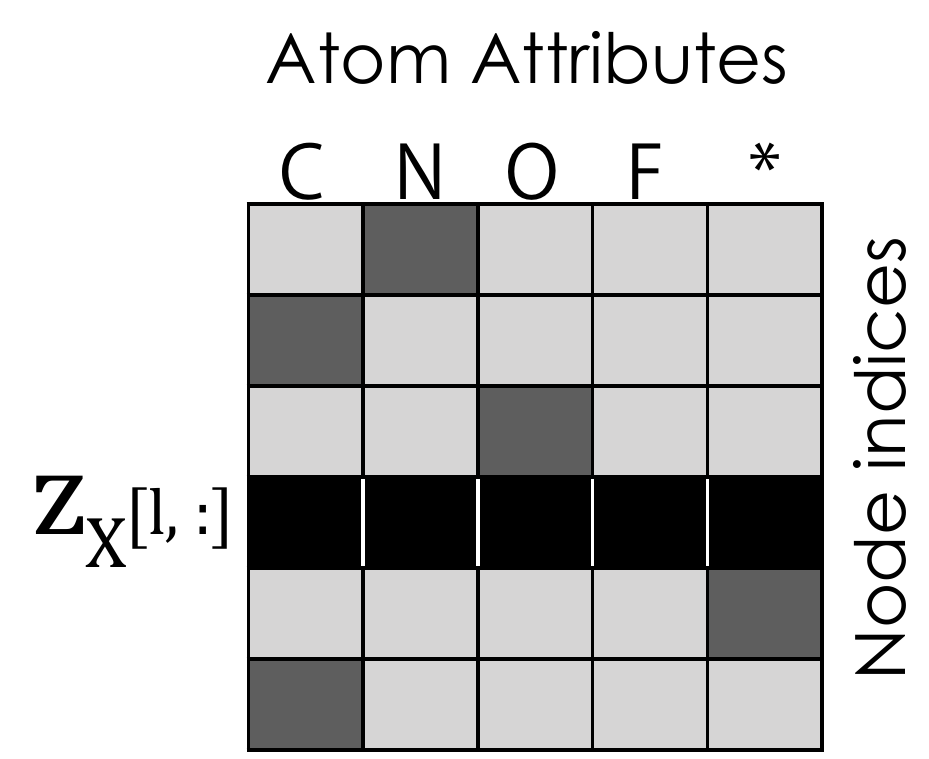}
		\caption{Node feature coupling layer: All channels belonging to a single node are masked.}
	\end{subfigure}
	\caption{Masking schemes used in proposed affine coupling layers.}
	\label{fig:coupling}
\end{figure}

Eqs. (\ref{eqn:node_feature_coupling},~\ref{eqn:adjacency_coupling}) are implemented with masking patterns shown in \autoref{fig:coupling}. Based on experimental evidence, we observe that masking $z_A (A')$ and $z_X (X')$ w.r.t. the node axis performs the best. 
Because a single coupling layer updates one single slice of $z_A$ and $z_X$, we need a sequence of $N$ coupling layers at the minimum, each masking a different node, for each of the adjacency coupling and the node feature coupling layers. 

We acknowledge that this choice of masking axis over $z_X$ and $z_A$ makes the transformations not invariant to permutations of the nodes. We can easily formulate permutation-invariant couplings by changing the slice indexing based on the non-node axes (the 3\textsuperscript{rd} axis of the adjacency tensor, and the 2\textsuperscript{nd} axis of the feature matrix). However, using such masking patterns results in dramatically worse performance due to the sparsity of molecular graphs. For example, organic compounds are mostly made of carbon atoms. Thus, masking the carbon column in $X'$ (and $z_X$) results in feeding a nearly-empty matrix to the scale and the translation networks, which is almost non-informative to update the carbon column entries of $X'$ and $z_X$. 
We consider this permutation dependency as a limitation of the current model, and we intend to work on this issue as future work.  

\subsection{Training}

During the training, we perform the forward computations shown in Figure \ref{fig:model_forward} over minibatches of training data ($G = (A, X)$) and obtain latent representations $z = \text{concat} (z_A, z_X)$. Our objective is maximizing the log likelihood (Eq.~\ref{eqn:log_likelihood}) of $z$ over minibatches of training data. This is implemented as minimization of the negative log likelihood using the Adam optimizer~\cite{Kingma_Ba15ICLR}.

\subsection{Two-step Molecular Graph Generation}

\begin{figure}[t]
    \centering
    \includegraphics[width=0.7\linewidth]{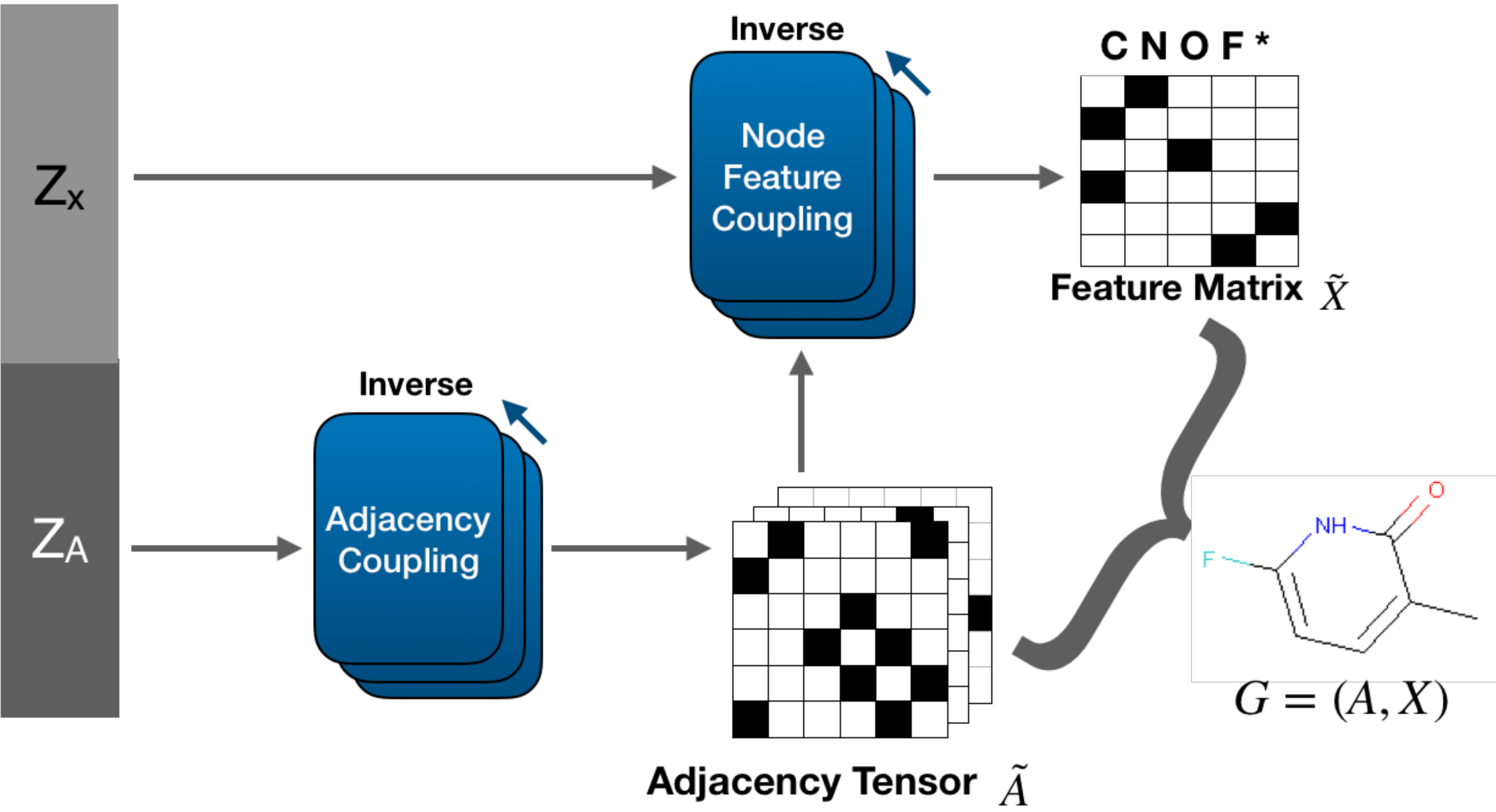}
    \caption{Generative process of the proposed GraphNVP. We apply the inverse of the coupling layers in the reverse order, so that the original input can be reconstructed.}
    \label{fig:model_reverse}
\end{figure}

Because our proposed model is invertible, graph generation is simply executing the process shown in Figure \ref{fig:model_forward} in reverse. During the training, node feature coupling and adjacency coupling can be performed in either order, as the output of one coupling module does not depend on the output of the other coupling module. However, because the node feature coupling module requires a valid adjacency tensor as an input, we also need an adjacency tensor to perform the reverse step of node feature coupling. 
Therefore, we apply the reverse step of adjacency coupling module first, so we get an adjacency tensor as the output. Next, the adjacency tensor is fed into the reverse step of the node feature coupling. The generation process is shown in Figure \ref{fig:model_reverse}. In section \ref{sec:experiments}, we show that this \textit{2-step generation process} can efficiently generate chemically valid molecular graphs. 

\textbf{1st step:} We draw a random sample $z = \text{concat} (z_A, z_X)$ from prior $p_z$ and split sampled $z$ into $z_A$ and $z_X$. 
Next, we apply a sequence of \textit{inverted} adjacency coupling layers on $z_A$. 
As a result, we obtain a probabilistic adjacency tensor $\tilde{A'}$, 
from which we construct a discrete adjacency tensor $\tilde{A} \in \{0, 1\}^{N\!\times N\!\times R}$ by taking node-wise and edge-wise argmax. 

\textbf{2nd step:} We generate a feature matrix given the sampled $z_X$ and the generated adjacency tensor $\tilde{A}$. 
We input $\tilde{A}$ along with $z_X$ into a sequence of \textit{inverted} node feature coupling layers to attain $\tilde{X'}$. Likewise, we take node-wise argmax of $\tilde{X'}$ to get discrete feature matrix $\tilde{X} \in \{0, 1\}^{N\!\times M}$. 

\section{Experiments} \label{sec:experiments}

\subsection{Procedure}

We use two popular chemical molecular datasets, QM9~\cite{ramakrishnan2014quantum}  and ZINC-250k~\cite{irwin2012zinc}. QM9 dataset contains 134k molecules, and ZINC-250k is made of 250k drug-like molecules randomly selected from the ZINC database. The maximum number of atoms in a molecule are 9 for the QM9 and 38 for the ZINC, respectively (excluding hydrogen). Following a standard procedure, we first kekulize molecules and then remove hydrogen atoms from them. The resulting molecules contain only single, double, and triple bonds. 

We convert each molecule to an adjacency tensor $A \in \{0, 1\}^{N\!\times\!N\!\times\!R}$ and a feature matrix $X \in \{0, 1\}^{N\!\times\!M}$. 
N is the maximum number of atoms a molecule in a certain dataset can have. If a molecule has less than $N$ atoms, we insert virtual nodes as padding to keep the dimensions of $A$ and $X$ the same for all the molecules. Because the original adjacency tensors can be sparse, we add a virtual bond edge between the atoms that do not have a bond in the molecule. Thus, an adjacency tensor consists of $R\!=\!4$ adjacency matrices stacked together, each corresponding to the existence of a certain type of bond (single, double, triple, and virtual bonds) between the atoms. The feature matrix is used to represent the type of each atom (e.g., oxygen, fluorine, etc.).

We use a multivariate Gaussian distribution $\mathcal{N}(\bm{0}, \sigma^2 \bm{I})$ as prior distribution $p_z(z)$, where standard deviation $\sigma$ is learned simultaneously during the training. 
We present more details in the appendix. 

\subsection{Numerical Evaluation}

Following~\cite{glow2018}, we sample 1,000 latent vectors from a temperature-truncated normal distribution $p_{z, T}(z)$ and transform them into molecular graphs by performing the reverse step of our model. We compare the performance of the proposed model with baseline models in \autoref{tab:results} using the following metrics.
\textbf{Validity (V)} is the percentage of generated graphs corresponding to valid molecules.
\textbf{Novelty (N)} is the percentage of generated valid molecules not present in the training set.
\textbf{Uniqueness (U)} is the percentage of unique valid molecules out of all generated molecules.
\textbf{Reconstruction accuracy (R)} is the percentage of molecules that can be reconstructed perfectly by the model: namely, the ratio of molecules $G$ s.t. $G = f^{-1}_{\theta} \left( f_{\theta} \left( G \right) \right)$. 

We choose Regularizing-VAE (RVAE)~\cite{ma2018constrained} and MolGAN~\cite{de2018molgan} as state-of-the-art baseline models. We compare with two additional VAE models; grammar VAE(GVAE)~\cite{kusner2017gvae} and 
character VAE (CVAE)\cite{gomez2018automatic}, which learn to generate string representations of molecules. In this paper, we do not conduct comparisons with some models that generate nodes and edges sequentially\cite{jtvae2018, cgvae2018}, because our model generates adjacency tensor in one shot. 


Notably, proposed GraphNVP guarantees 100\% reconstruction accuracy, attributed to the invertible function construction of normalizing flows. Also, it is notable that GraphNVP enjoys a significantly high uniqueness ratio. Although some baselines exhibit a higher validity on QM9 dataset, the set of generated molecules contains many duplicates. 
Additionally, we want to emphasize that our model generates a substantial number of valid molecules without explicitly incorporating the chemical knowledge as done in some baselines (e.g., valency checks for chemical graphs in MolGAN and RVAE). This is preferable because additional validity checks consume computational time, and may result in a low reconstruction accuracy (e.g., RVAE). As GraphNVP does not incorporate domain-specific procedures during learning, it can be easily used for learning generative models on general graph structures.

\begin{table}[t]
	\centering
    {\tabcolsep = 1.5mm
	\begin{tabular}{l|cccc|cccc}
		\multirow{2}{*}{Method}  & \multicolumn{4}{c|}{QM9} & \multicolumn{4}{c}{ZINC} \\
		& \% V & \% N & \% U & \% R & \% V & \% N & \% U & \% R \\ \hline
		\textbf{GraphNVP} & 
		\begin{tabular}{c} 83.1 \\ \footnotesize{($\pm$ 0.5)} \end{tabular} &
		\begin{tabular}{c} 58.2 \\ \footnotesize{($\pm$ 1.9)} \end{tabular} & 
		\begin{tabular}{c} 99.2 \\ \footnotesize{($\pm$ 0.3)} \end{tabular} & 100.0 
		& \begin{tabular}{c} 42.6 \\ \footnotesize{($\pm$ 1.6)} \end{tabular} &
		\begin{tabular}{c} 100.0 \\ \footnotesize{($\pm$ 0.0)} \end{tabular} & 
		\begin{tabular}{c} 94.8 \\ \footnotesize{($\pm$ 0.6)} \end{tabular} & 100.0\\
		RVAE~\cite{ma2018constrained} & 96.6 & 97.5  & - & 61.8 & 34.9  & 100.0  & - & 54.7\\
		MolGAN~\cite{de2018molgan} & 98.1 & 94.2 & 10.4 & - & - & - & - & - \\
		GVAE~\cite{kusner2017gvae} & 60.2 & 80.9 & 9.3 & 96.0 & 7.2 & 100.0 & 9.0 & 53.7\\
		CVAE~\cite{gomez2018automatic} & 10.3 & 90.0 & 67.5 & 3.6 & 0.7 & 100.0 & 67.5 & 44.6
	\end{tabular}
	}
    \vspace*{2mm}
	\caption{Performance of generative models with respect to quality metrics. Baseline scores are borrowed from the original papers. Scores of GraphNVP are averages over 5 runs. Standard deviations are presented below the average scores.}
	\label{tab:results}
\end{table}

Remarkably, GraphNVP could achieve a high uniqueness score without incorporating domain expert knowledge as done in previous work~\cite{ma2018constrained}. Next, we provide a brief discussion of our findings. 

The authors of MolGAN~\cite{de2018molgan} report that MolGAN is susceptible to \textit{mode collapse}~\cite{salimans2016improved} resulting in a very small number of unique molecules among generated molecules. We found this as a reasonable observation, explaining the score in Table~\ref{tab:results}. 

Usually we expect VAEs to find a latent space serving as a low-dimensional and smooth approximation of the true sample distribution. The VAE latent spaces are thus expected to \textit{omit} some minor variations in samples to conform to a smooth low-dimensional latent distribution. 
In contrast, in the latent space of invertible flow models, \textit{no omissions of minor graph variations are allowed} since all encodings (forwarding) must be analytically invertible. And the latent space of the invertible flows are not low-dimensional: basically, the dimensionality does not change during affine transformations. We conjecture that these distinctions of latent spaces may partially explain the observed difference of uniqueness among two types of models.

\subsection{Smoothness of the Learned Latent Space}

\begin{figure}[t]
	\centering
	\vspace{-0cm}
	\begin{subfigure}[b]{0.49\textwidth}
		\begin{center}
			\centering
			\includegraphics[width=1.0\textwidth]{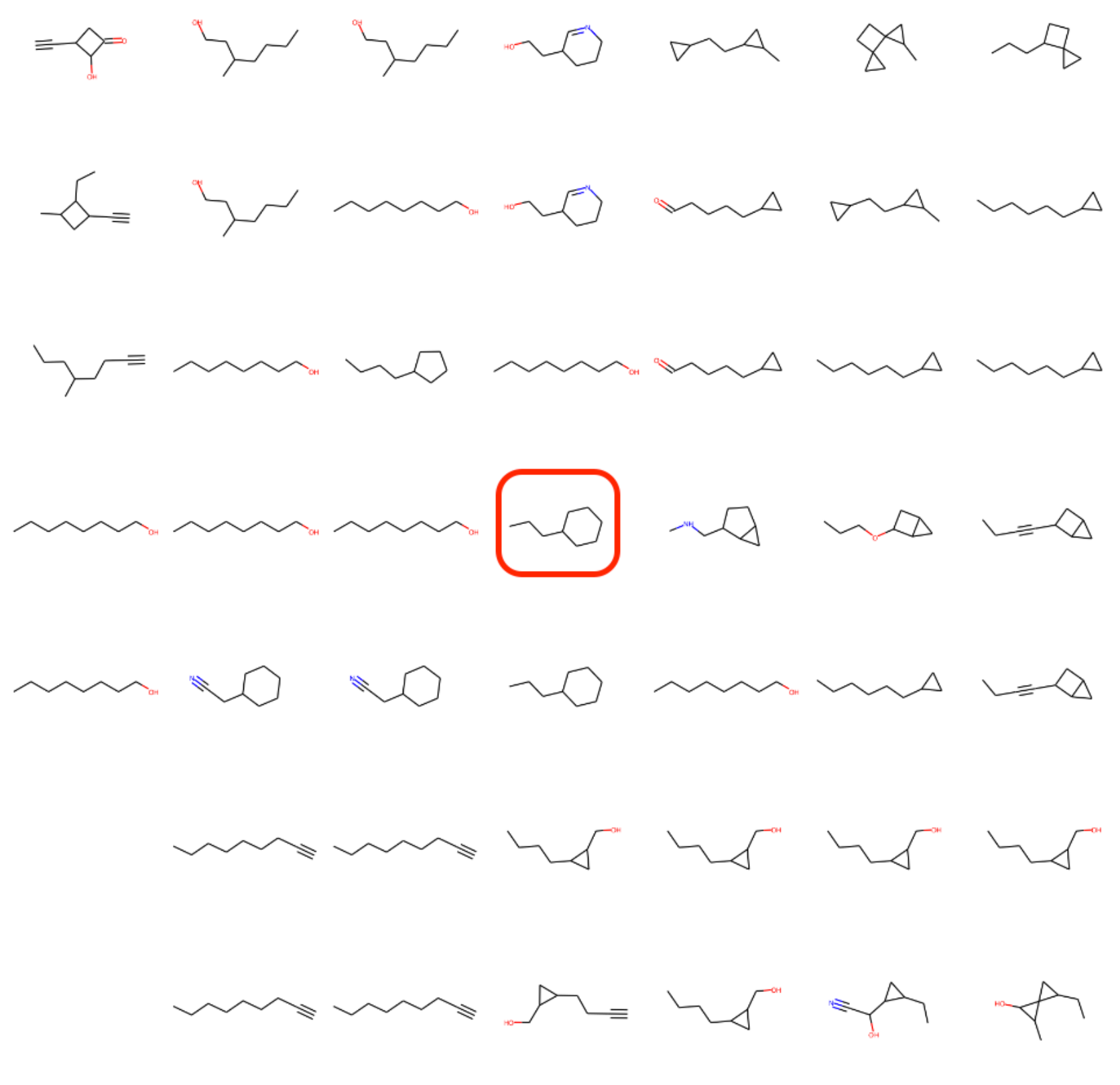}
			\caption{Learned latent space for QM9.}
		\end{center}
	\end{subfigure}%
	\hspace{1mm}
	\begin{subfigure}[b]{0.49\textwidth}
		\centering
		\includegraphics[width=1.0\textwidth]{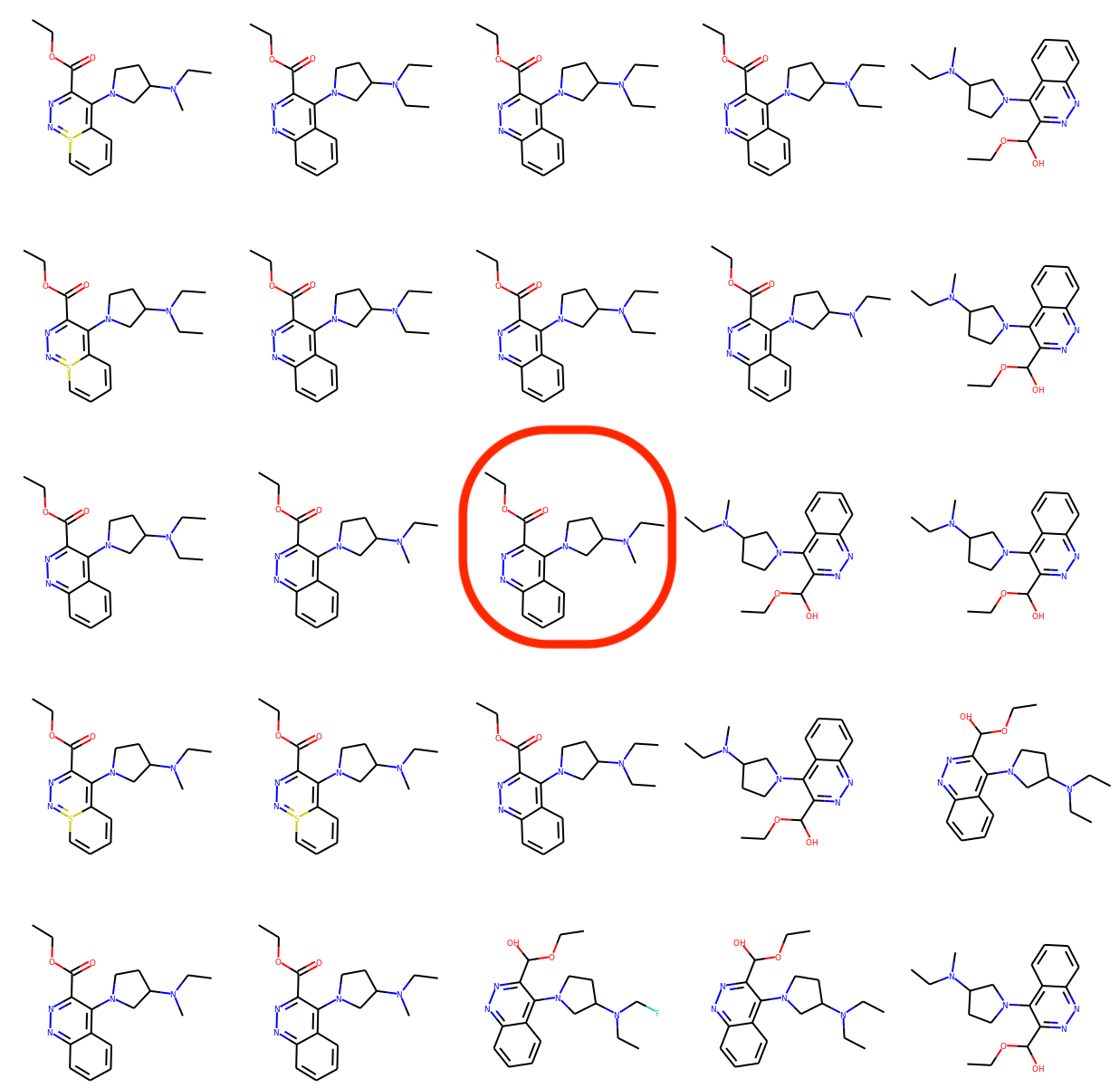}
		\caption{Learned latent space for ZINC.}
	\end{subfigure}
	\caption{Visualization of the learned latent spaces along two randomly selected orthogonal axes. The red circled molecules are centers of the visualizations (not the origin of the latent spaces). An empty space in the grid indicates that an invalid molecule is generated.}
	\label{fig:neighborhood_2d}
\end{figure}

Next, we qualitatively examine the learned latent space $z$ by visualizing the latent points space. In this experiment, we randomly select a molecule from the training set and encode it into a latent vector $z_0$ using our proposed model. Then we choose two random axes which are orthogonal to each other. We decode latent points lying on a 2-dimensional grid spanned by those two axes and with $z_0$ as the origin. \autoref{fig:neighborhood_2d} shows that the latent spaces learned from both QM9 (panel (a)) and ZINC dataset (panel (b)) vary smoothly such that neighboring latent points correspond to molecules with minor variations. This visualization indicates the smoothness of the learned latent space, similar to the results of existing VAE-based models (e.g.,~\cite{cgvae2018,ma2018constrained}). However, it should be noted that we decode each latent point only once unlike VAE-based models. For example, GVAE~\cite{kusner2017gvae} decodes each latent point 1000 times and selects the most common molecule as the representative molecule for that point. Because our decoding step is deterministic such a time-consuming measure is not needed. In practice, smoothness of the latent space is crucial for \textit{decorating} a molecule: generating a slightly-modified graph by perturbing the latent representation of the source molecular graph. 

\subsection{Property-Targeted Molecule Optimization}

\begin{figure}[t]
	\centering
	\vspace{-0cm}
	\begin{subfigure}[b]{0.9\textwidth}
		\begin{center}
			\centering
			\includegraphics[width=0.85\textwidth]{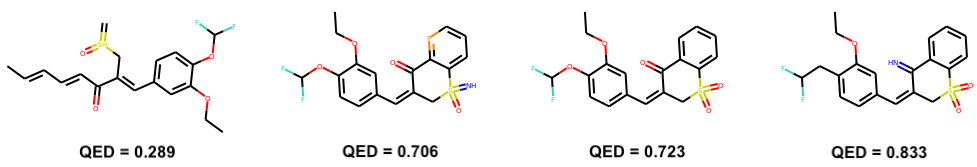}
			\caption{Molecule optimization for ZINC.}
		\end{center}
	\end{subfigure}%
	\hspace{1mm}
	\begin{subfigure}[b]{1.0\textwidth}
		\centering
		\includegraphics[width=0.65\textwidth]{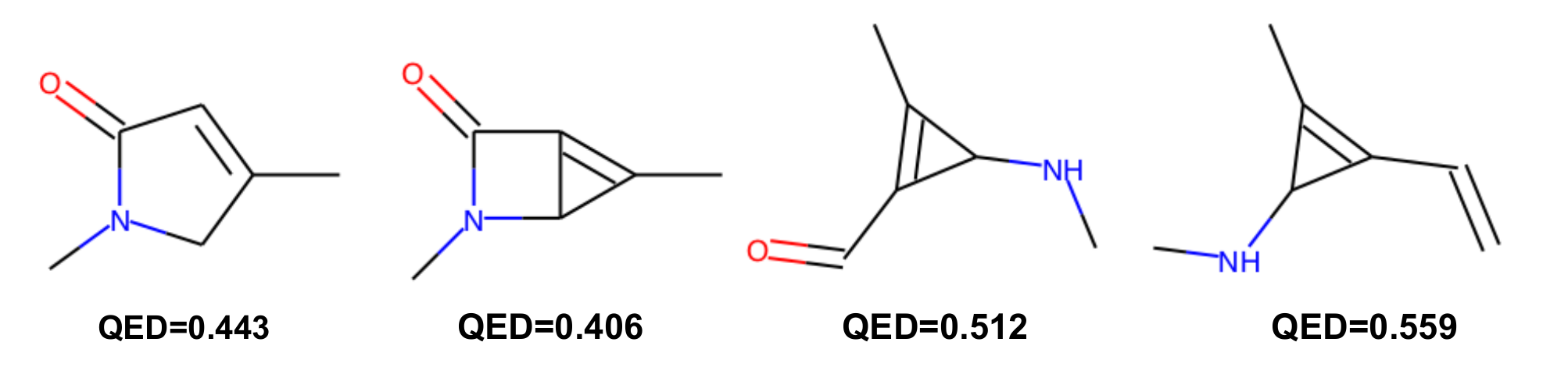}
		\caption{Molecule optimization for QM9.}
	\end{subfigure}
	\caption{Chemical property optimization. Given the left-most molecule, we interpolate its latent vector along the direction which maximizes its QED property.}
	\label{fig:ZINC_propoerty}
\end{figure}

Our last task is to find molecules similar to a given molecule, but possessing a better chemical property. This task is known as \textit{molecular optimization} in the field of chemo-informatics. We train a linear regressor on the latent space of molecules with quantitative estimate of drug-likeness (QED) of each molecule as the target chemical property. QED score quantifies how likely a molecule is to be a potential drug. We interpolate the latent vector of a randomly selected molecule along the direction of increasing QED score as learned by linear regression. 
\autoref{fig:ZINC_propoerty} demonstrates the learned latent space and a simple linear regression yields successful molecular optimization. Here, we select a molecule with a low QED score and visualize its neighborhood. However, we note that the number of valid molecules that can be generated along a given direction varies depending on the query molecule.
We show another property optimization example on QM9 dataset in the appendix. 

Although we could perform molecular optimization with linear regression, we believe an extensive Bayesian optimization (e.g., ~\cite{jtvae2018, kusner2017gvae}) on the latent space may provide better results.

\section{Conclusion}

In this paper, we propose GraphNVP, an invertible flow-based model for generating molecular graphs, first in the literature. Our model can generate valid molecules with a high uniqueness score and guaranteed reconstruction ability. In addition, we demonstrate that the learned latent space can be used to search for molecules similar to a given molecule, which maximizes a desired chemical property. 
There is an important open problem: how to improve the permutation-invariance of the proposed model. Additionally, we believe more exploration of the reasons contributing to the high uniqueness ratio of the proposed model will contribute to the understanding of graph generation models in general. 

\bibliography{neurips}
\bibliographystyle{plain}

\newpage
\appendix
\section{Network Architecture details}

For QM9 dataset, we use a total of 27 adjacency coupling and 36 node feature coupling layers. For ZINC dataset, we keep the number of coupling layers equal to the maximum number of atoms a ZINC molecule can have, 38. We model affine transformation (both scale and translation) of an adjacency coupling layer with a multi-layer perceptron (MLP). As mentioned in the main text, we utilize both node assignments and adjacency information in defining node feature coupling layers. However, we found affine transformations can become unstable when used to update the feature matrix with Relational-GCN (RelGCN). Therefore, we use only additive transformations in node feature coupling layers.

We initialize the last layer of each RelGCN and MLP with zeros, such that each affine transformation initially performs an identity function. 

We train the models using Adam optimizer with default parameters ($\alpha$ = 0.001) and minibatch sizes 256 and 128 for QM9 and ZINC datasets. We use batch normalization in both types of coupling layers.  

\section{Training Details}
For training data splits, we used the same train/test dataset splits used in~\cite{kusner2017gvae}.
We train each model for 200 epochs. We did not employ early-stopping in the experiments. We chose the model snapshot of the last (200) epoch for evaluations and demonstrations.
All models are implemented using Chainer-Chemistry\footnote{https://github.com/pfnet-research/chainer-chemistry} and RDKit\footnote{https://github.com/rdkit/rdkit} libraries. 

\section{Effect of temperature}
\label{appendix:temp}
\begin{figure} [htbp]
	\centering
	\vspace{-0cm}
	\begin{subfigure}[b]{0.49\textwidth}
		\begin{center}
			\centering
			\includegraphics[width=1.0\textwidth]{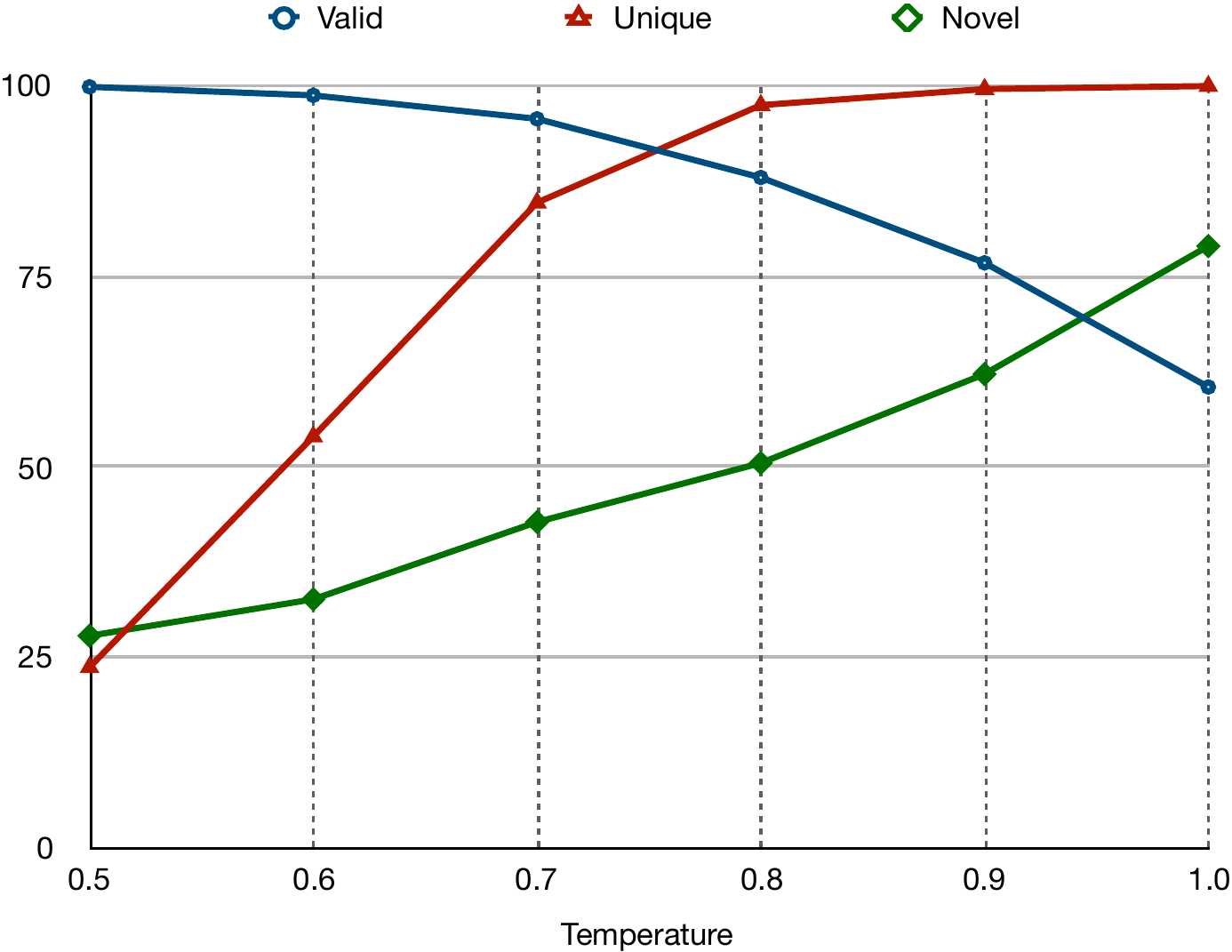}
			\caption{Impact of temperature on sampling from latent space of QM9.}
		\end{center}
	\end{subfigure}%
	\hspace{1mm}
	\begin{subfigure}[b]{0.49\textwidth}
		\centering
		\includegraphics[width=1.0\textwidth]{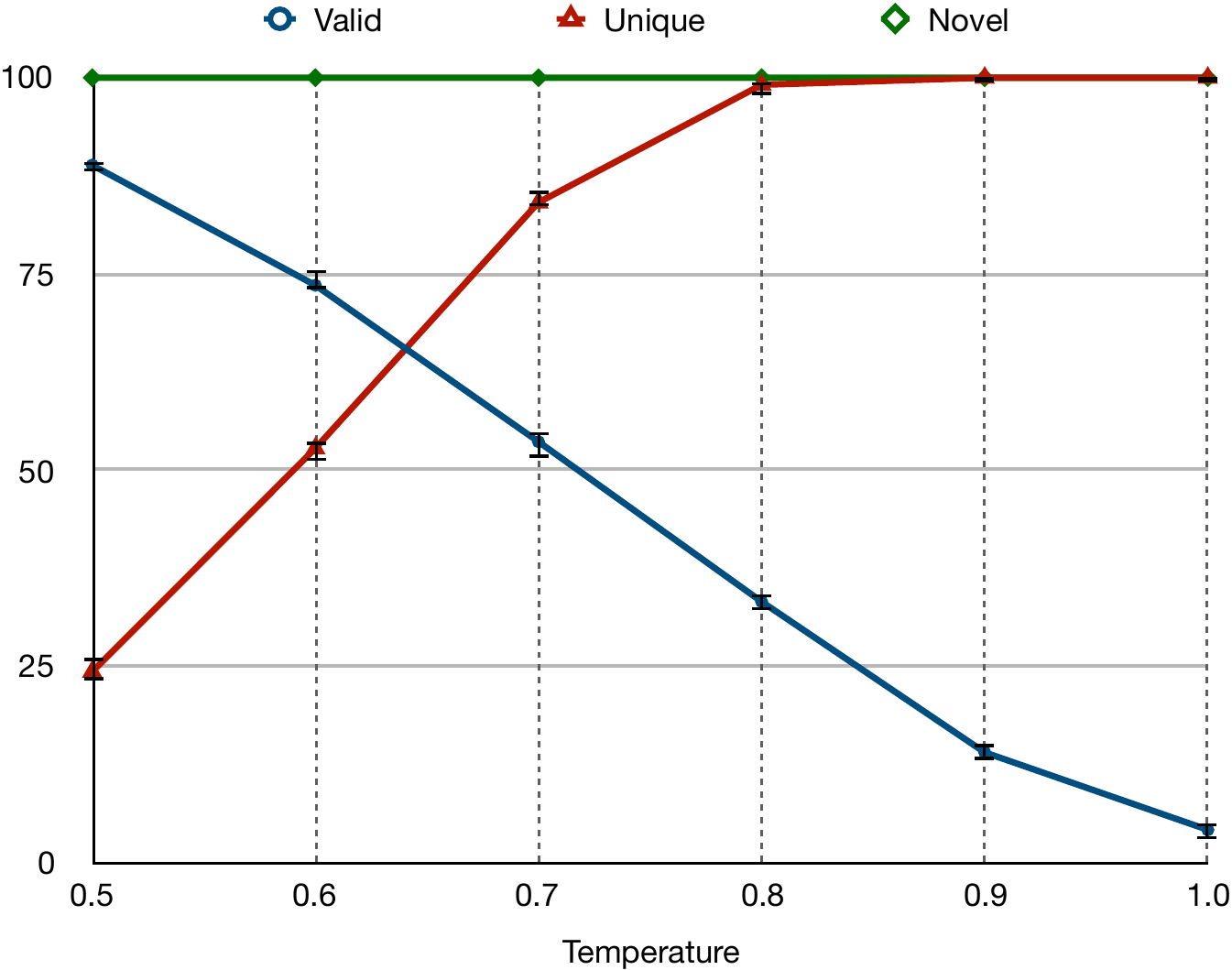}
		\caption{Impact of temperature on sampling from latent space of ZINC.}
	\end{subfigure}
	\caption{Impact of temperature on the quality of graph generation. Sampling with a smaller temperature yields more valid molecules but with less diversity (uniqueness) among them. Each experiment is performed five times and the average is reported in this figure.}
	\label{fig:temp_sampling}
\end{figure}

Following previous work on likelihood-based generative models \cite{glow2018}, we sampled latent vectors from a temperature-truncated normal distribution. Sampling with a lower temperature results in higher number of valid molecules at the cost of uniqueness among them. How temperature effects validity, uniqueness, and novelty of generated molecules is shown in Figure \ref{fig:temp_sampling}. Based on empirical results we chose 0.85 and 0.75 as the temperature values for QM9 and ZINC models respectively.

\section{Additional Visualizations} \label{sec:additional_visualization}

\begin{figure} [h]
    \centering
    \includegraphics[width=0.8 \linewidth]{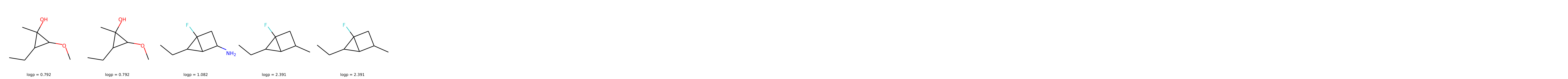}
    \caption{Chemical property optimization. We select a molecule from QM9 dataset randomly and then interpolate its latent vector along the axis which maximizes water-octanol partition coefficient (logP)}
    \label{fig:logp_interpolation}
\end{figure}

Fig.~\ref{fig:logp_interpolation} illustrates an example of chemical property optimization for water-octanol partition coefficient (logP) on QM9 dataset. 

\end{document}